\documentclass{article}

\usepackage{microtype}
\usepackage{graphicx}
\usepackage{subfigure}
\usepackage{booktabs} 
\usepackage{natbib}
\usepackage{todonotes}
\usepackage{tabularx}
\usepackage{multicol}

\usepackage{amsfonts, amsmath}
\usepackage{bm}

\usepackage{hyperref}
\usepackage[capitalise]{cleveref}

\usepackage[accepted]{icml2021}\usepackage[accepted]{icml2021}

\icmltitlerunning{Contrastive Predictive Coding for Anomaly Detection and Segmentation}

\begin{document}

\twocolumn[
\icmltitle{Contrastive Predictive Coding for Anomaly Detection and Segmentation}




\begin{icmlauthorlist}
\icmlauthor{Puck de Haan}{to}
\icmlauthor{Sindy L\"owe}{to}
\end{icmlauthorlist}

\icmlaffiliation{to}{UvA-Bosch Delta Lab, University of Amsterdam, Netherlands}

\icmlcorrespondingauthor{Puck de Haan}{pdehaan274@gmail.com}
\icmlcorrespondingauthor{Sindy L\"owe}{loewe.sindy@gmail.com}

\icmlkeywords{Machine Learning, ICML, Anomaly Detection, Contrastive Learning}

\vskip 0.3in
]



\printAffiliationsAndNotice{}  

\begin{abstract} 
 Reliable detection of anomalies is crucial when deploying machine learning models in practice, but remains challenging due to the lack of labeled data. To tackle this challenge, contrastive learning approaches are becoming increasingly popular, given the impressive results they have achieved in self-supervised representation learning settings.  However, while most existing contrastive anomaly detection and segmentation approaches have been applied to images, none of them can use the contrastive losses directly for both anomaly detection and segmentation. In this paper, we close this gap by making use of the Contrastive Predictive Coding model \citep{oord2018}. We show that its patch-wise contrastive loss can directly be interpreted as an anomaly score, and how this allows for the creation of anomaly segmentation masks.
 The resulting model achieves promising results for both anomaly detection and segmentation on the challenging MVTec-AD dataset.
\end{abstract}

\section{Introduction}
An anomaly (or outlier, novelty, out-of-distribution sample) is an observation that differs significantly from the vast majority of the data. Anomaly detection (AD) tries to distinguish anomalous samples from the samples that are deemed `normal' in the data. It has become increasingly relevant to detect these anomalies to make machine learning methods more reliable and to improve their applicability in real-world scenarios, such as automated industrial inspections and medical diagnosis \cite{ruffoverview2021}. Typically, anomaly detection is treated as an unsupervised learning problem, since labelled data is generally unavailable and to allow for the development of methods that can detect previously unseen anomalies.

One promising direction involves the adaptation of contrastive learning approaches \cite{hjelm2018learning, oord2018, chen2020simple, He_2020_CVPR} to the anomaly detection setting \cite{tack2020csi, winkens2020contrastive, Kopuklu_2021_WACV, qiu2021neural, sohn2021learning}. However, even though most of these approaches have been applied to image data, none of them can use the contrastive losses directly for both anomaly detection and segmentation.

In this paper, we demonstrate that Contrastive Predictive Coding (CPC) \cite{oord2018, henaff2020dataefficient} can be applied to detect and segment anomalies in images. We show that the InfoNCE loss introduced by \citet{oord2018} can be directly interpreted as an anomaly score. Since in this loss patches from within an image are contrasted against one another, we can further use it to create accurate anomaly segmentation masks. This results in a compact and straightforward anomaly detection and segmentation approach.


\begin{figure}[t]
    \centering
    \includegraphics[width=0.55\textwidth, keepaspectratio]{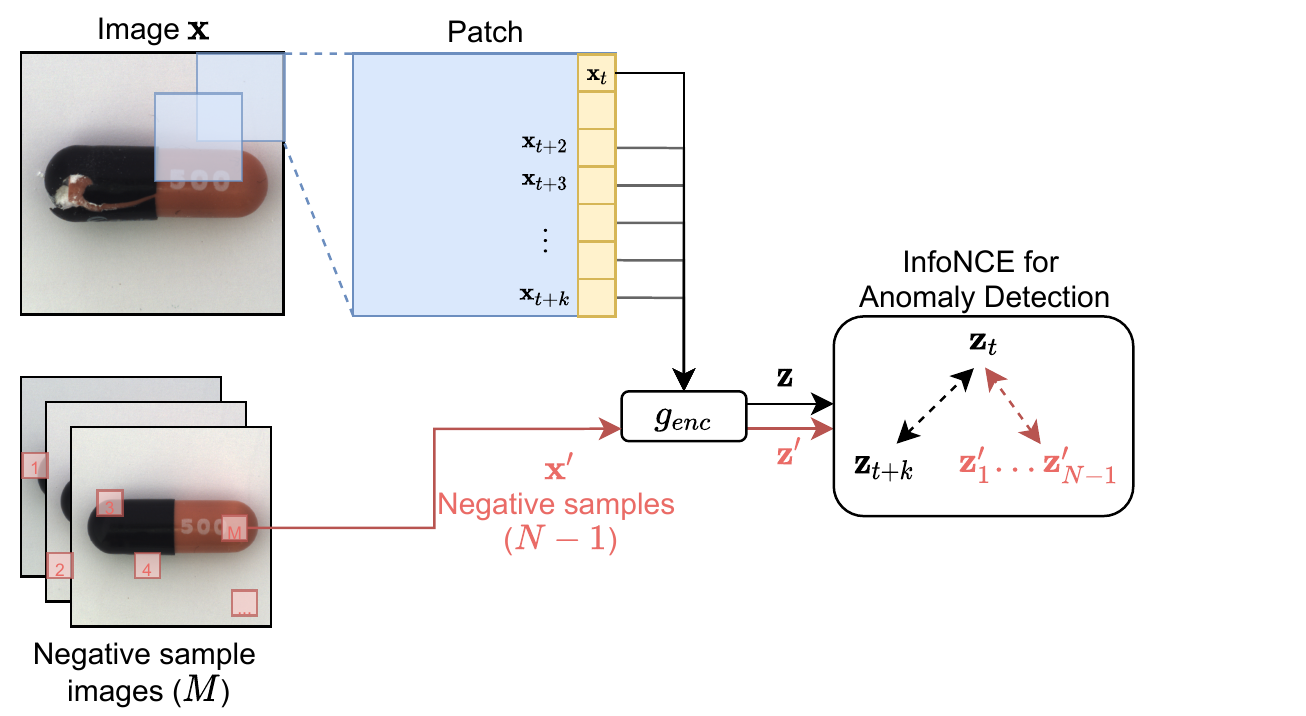}
    \vspace{-3mm}
    \caption{A schematic overview of Contrastive Predictive Coding for anomaly detection and segmentation in images. After extracting (sub-)patches from the input image, we contrast the encoded representations from within the same image $(\bm{z}_{t}, \bm{z}_{t+k})$ against $N-1$ randomly matched representations $(\bm{z}_{t}, \bm{z}_j)$. The resulting InfoNCE loss is used to determine whether sub-patch $\bm{x}_{t+k}$ is anomalous or not.}
    \vspace{-5mm}
    \label{fig:model}
\end{figure}

To improve the performance of the CPC model for anomaly detection, we introduce two adjustments.
First, we adapt the setup of negative samples during testing such that anomalous patches can only appear within the positive sample. 
Second, we omit the autoregressive part of the CPC model. 
With these adjustments, our proposed method achieves promising performance on real-world data, such as the challenging MVTec-AD dataset \cite{Bergmann_2019_CVPR}.

\section{Related Work} 
In this section, we will give an overview of contrastive learning approaches and different methods for anomaly detection. 

\subsection{Contrastive Learning}
Lately, impressive results have been achieved with self-supervised methods based on contrastive learning \cite{Wu_2018_CVPR, oord2018, hjelm2018learning, He_2020_CVPR, chen2020simple, li2021prototypical}. Overall, these methods work by making a model decide whether two (randomly) transformed inputs originated from the same input sample, or from two samples that have been randomly drawn from across the dataset. Different transformations can be chosen depending on the domain and downstream task. For example, on image data, random data augmentation such as random cropping and color jittering has proven useful \citep{chen2020simple, He_2020_CVPR}. In this paper, we use the Contrastive Predictive Coding model \citep{oord2018, henaff2020dataefficient}, which makes use of temporal transformations. Generally, these approaches are evaluated by training a linear classifier on top of the created representations and by measuring the performance that this linear classifier can achieve on downstream tasks.

\subsection{Anomaly Detection}
Anomaly detection methods can roughly be divided into three categories: density-based, reconstruction-based and discriminative-based methods \cite{ruffoverview2021}. Density-based methods predict anomalies by estimating the probability distribution of the data (e.g. GANs, VAEs, or flow-based models) \cite{schlegl2017unsupervised, winkens2020contrastive, Liu_2020_CVPR}; reconstruction-based methods are based on models that are trained with a reconstruction objective (e.g. autoencoders) \cite{chong2017, bergmann2018improving, luo2020encoding}; discriminative-based methods learn a decision boundary between anomalous and normal data (e.g. SVM, one-class classification) \cite{ruff2020deep,  tack2020csi, liznerski2020explainable, li2021cutpaste}. The method proposed in this paper can be seen as a density-based method with a discriminative one-class objective. 

Several previous works investigate the use of contrastive learning for AD. \citet{tack2020csi, winkens2020contrastive,sohn2021learning} make use of the SimCLR framework \cite{chen2020simple} to learn representations of the data. Then, they calculate a separate anomaly score by using these representations for density estimation, one-class classification, or by applying metric measures like the cosine similarity and the norm of the representations. The downsides of this approach are that it requires extensive data augmentations and multiple different measures, or multiple models. Another comparable contrastive learning AD method \cite{Kopuklu_2021_WACV} uses noise contrastive estimation for training, similar to our method. Differently to our method, they map the samples to multiple latent spaces and use anomalous samples as negatives during training. This results in a more complex model with a supervised training phase. NeuTraL AD \cite{qiu2021neural} makes use of a contrastive loss with learnable transformations, and reuses this loss as an anomaly score. In contrast to our method, their approach has been evaluated on time-series and tabular data.

\begin{figure*}[t]
    \centering
    \includegraphics[width=0.95\textwidth]{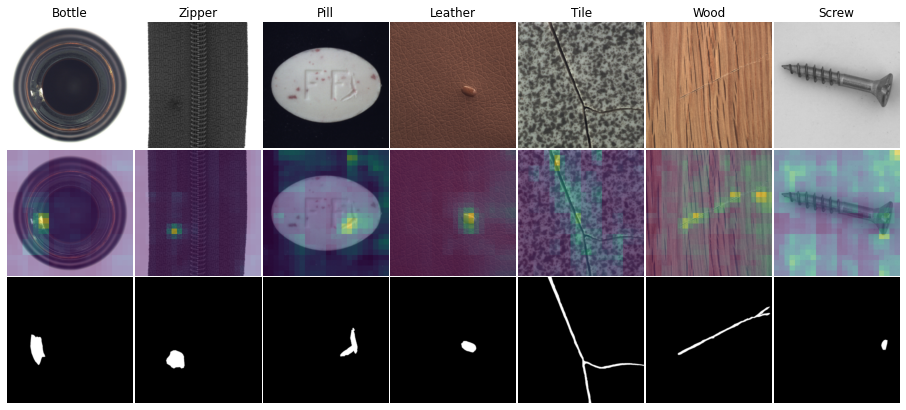}
    \caption{Localization of anomalous regions for different classes in the MVTec-AD dataset. The top row shows the original input image, the mid row depicts the superimposition of the image and corresponding InfoNCE loss values (brighter colors represent higher loss values) and the bottom row shows the ground truth annotation. We find that our model consistently highlights anomalous regions across many classes. One notable exception is the screw class (right), for which the model assigns high loss values to the background in many cases.}
    \label{fig:segmentation_results}
\end{figure*}

\section{Contrastive Predictive Coding}
Contrastive Predictive Coding \cite{oord2018} is a self-supervised representation learning approach that leverages the structure of the data and enforces temporally nearby inputs to be encoded similarly in latent space. It achieves this by making the model decide whether a pair of samples is made up of temporally nearby samples or randomly assigned samples. 
This approach can also be applied to static image data by splitting the images up into patches, and interpreting each row of patches as a separate time-step.

The CPC model makes use of a contrastive loss function, coined InfoNCE, that is based on Noise-Contrastive Estimation \cite{gutmann10a} and is designed to optimize the mutual information between the latent representations of patches ($\bm{z}_t$) and their surrounding patches ($\bm{c}_{t+k}$):
\begin{equation}
    \mathcal{L}_k = -\mathbb{E}_X \left[ \log\frac{\exp(\bm{z}_{t+k}^\top\bm{W}_k \bm{c}_t)}{\sum_X\exp(\bm{z}_j^\top \bm{W}_k \bm{c}_t)}\right] ~~~,
    \label{eq:infonce}
\end{equation}
where $\bm{z}_t = g_{\text{enc}}(\bm{x}_t)$ and $g_{\text{enc}}$ represents a non-linear encoder, $\bm{c}_t = g_{\text{ar}}(\bm{z}_{\leq t})$ and $g_{\text{ar}}$ represents an autoregressive model. Furthermore, $\bm{W}_k$ describes a linear transformation used for predicting $k$ time-steps ahead. The set of samples $X$ consists of one positive sample $(\bm{x}_{t}, \bm{x}_{t+k})$ and $N-1$ negative samples $(\bm{x}_{t}, \bm{x}_j)$ for which $\bm{x}_j$ is randomly sampled from across the current batch. 

\section{CPC for Anomaly Detection}
We propose to apply the CPC model for anomaly detection and segmentation (\cref{fig:model}). 
In order to improve the performance of the CPC model in this setting, we introduce two adjustments to its architecture: 
\textbf{(1)} We omit the autoregressive model $g_{\text{ar}}$. As a result, our loss function changes to:
\begin{equation}
    \mathcal{L}_k = -\mathbb{E}_X \left[ \log\frac{\exp(\bm{z}_{t+k}^\top\bm{W}_k \bm{z}_t)}{\sum_X\exp(\bm{z}_j^\top \bm{W}_k \bm{z}_t)}\right] ~~~.
    \label{eq:new_infonce}
\end{equation}
This formulation is equivalent to the loss used in the Greedy InfoMax model \cite{lowe2019putting}. This adjustment results in a simpler model, which is still able to learn useful latent representations -- according to preliminary results.
\textbf{(2)} We change the setup of the negative samples during testing. Previous implementations of the CPC model use random patches from within the same test-batch \cite{henaff2020dataefficient} as negative samples. However, this may result in negative samples containing anomalous patches, which could make it harder for the model to detect anomalous patches in the positive sample based on the contrastive loss. To avoid this, during testing, we draw negative samples from the (non-anomalous) training data.

In the test-phase, we use the loss function in \cref{eq:new_infonce} to decide whether an image patch $\bm{x}_{t+k}$ can be classified as anomalous:
\begin{equation}
    \bm{x}_{t+k} = 
    \begin{cases}
        \text{anomalous,~~~~~ if } \mathcal{L}_k \geq \tau\\
        \text{normal,~~~~~~~~~~~ if } \mathcal{L}_k < \tau ~~~.
    \end{cases}
\end{equation}
The threshold value $\tau$ remains implicit, since we use the area under the receiver operating characteristic curve (AUROC) as performance measure. While we can create anomaly segmentation masks by making use of the anomaly scores per patch $\bm{x}_{t+k}$, we can also apply our approach to decide whether a sample is anomalous -- either by averaging over the scores of all patches within an image, or by examining the patch with the highest score.


\section{Experiments}
We evaluate the proposed Contrastive Predictive Coding model for anomaly detection and segmentation on the MVTec-AD dataset \cite{Bergmann_2019_CVPR}. This dataset contains high-resolution images of ten objects and five textures with pixel-accurate annotations and provides between 60 and 391 training images per class. During training we randomly crop every image to $[0.8-1]$ times the original dimensions. Then, both train and test images are resized to 768$\times$768 pixels. The resulting image is split into patches of size 256$\times$256, where each patch has 50\% overlap with its neighbouring patches. These patches are further divided into sub-patches of size 64$\times$64, also with 50\% overlap. These sub-patches are used in the InfoNCE loss (\cref{fig:model}) to detect anomalies. The cropped and resized images are horizontally flipped with a probability of 50\% during training. 

We use a ResNet-18 v2 \cite{he2016identity} up until the third residual block as encoder $g_{\text{enc}}$. We train a separate model from scratch for each class with a batch size of 16 for 150 epochs using the Adam optimizer \citep{kingma2014adam} with a learning rate of $1.5\mathrm{e}{-4}$. As proposed by \citet{oord2018}, we train and evaluate the model on grayscale images. For both training and evaluation we use 16 negative samples in the InfoNCE loss. To increase the accuracy of the InfoNCE loss as an indicator for anomalous patches, we apply four separate models in four different directions -- predicting patches using context from above, below, left and right -- and combine their losses in the test-phase.

\subsection{Anomaly Detection}

To evaluate our model's performance for detecting anomalies, we average the top-$5\%$ InfoNCE loss values across all sub-patches within an image and use this value to calculate the AUROC score. 
In \cref{tab:auc_detection}, we compare against previously published works from peer-reviewed venues that do not make use of pre-trained feature extractors. We find that our proposed \textbf{CPC-AD} model substantially improves upon a kernel density estimation model (\textbf{KDE}) and an autoencoding model (\textbf{Auto}) as presented in \citet{kauffmann2020clever}. We also improve upon the contrastive learning approach combined with a KDE model (avg. AUROC: 0.865) as proposed by \citet{sohn2021learning}. The performance of our model lags behind the \textbf{CutPaste} model \citep{li2021cutpaste}. However, we argue that CPC-AD provides a more generally applicable approach for anomaly detection. The CutPaste model relies heavily on randomly sampled artificial anomalies that are designed to resemble the anomalies encountered in the dataset. As a result, it is not applicable to a $k$-classes-out task, where anomalies differ semantically from the normal data. 
For comparison, the current state-of-the-art model on this dataset which makes use of a pre-trained feature extractor achieves 0.979 AUROC averaged across all classes \citep{defard2020padim}.

\begin{table}[t]
    \centering
    \caption{Anomaly detection AUROC score on the MVTec-AD test-set per category. We find that the proposed \textbf{CPC-AD} approach substantially outperforms the kernel density estimation model (\textbf{KDE}) and the autoencoding model (\textbf{Auto}) presented by \citet{kauffmann2020clever}. It is outperformed by the \textbf{CutPaste} model \citep{li2021cutpaste}, which relies heavily on dataset-specific augmentations for its training.}
    \begin{tabular}{rcccc} \toprule
        & \textbf{KDE} & \textbf{Auto} & \textbf{CutPaste} & \textbf{CPC-AD}   \\ \toprule
        Bottle     & 0.833 & 0.950 & 0.983 & 0.998 \\
        Cable      & 0.669 & 0.573 & 0.806 & 0.880 \\
        Capsule    & 0.562 & 0.525 & 0.962 & 0.641 \\
        Carpet     & 0.348 & 0.368 & 0.931 & 0.809 \\
        Grid       & 0.717 & 0.746 & 0.999 & 0.983 \\
        Hazelnut   & 0.699 & 0.905 & 0.973 & 0.996 \\
        Leather    & 0.415 & 0.640 & 1.000 & 0.990 \\
        Metal nut  & 0.333 & 0.455 & 0.993 & 0.845 \\
        Pill       & 0.691 & 0.760 & 0.924 & 0.921 \\
        Screw      & 0.369 & 0.779 & 0.863 & 0.897 \\
        Tile       & 0.689 & 0.518 & 0.934 & 0.957 \\
        Toothbrush & 0.933 & 0.494 & 0.983 & 0.878 \\
        Transistor & 0.724 & 0.512 & 0.955 & 0.925 \\
        Wood       & 0.947 & 0.885 & 0.986 & 0.803 \\
        Zipper     & 0.614 & 0.350 & 0.994 & 0.993 \\ \midrule
        Mean       & 0.636 & 0.631 & 0.952 & 0.901 \\ \bottomrule
    \end{tabular}
    \label{tab:auc_detection}
\end{table}

\subsection{Anomaly Segmentation}
For the evaluation of the proposed CPC-AD model's anomaly segmentation performance, we up-sample the sub-patch-wise InfoNCE loss values to match the pixel-wise ground truth annotations. To do so, we average the InfoNCE losses of overlapping sub-patches and assign the resulting values to all affected pixels. This allows us to create anomaly segmentation masks at the resolution of half a sub-patch (32$\times$32 pixels) that are of the same dimensions as the resized images (768$\times$768). 

In \cref{tab:segmentation_auc} in the Appendix, we compare the anomaly segmentation performance of the proposed CPC-AD method against previously published works from peer-reviewed venues.
The best results on the MVTec-AD dataset are achieved with extensive models that are pre-trained on ImageNet, such as \textbf{FCDD} and \textbf{PaDiM} \citep{liznerski2020explainable, defard2020padim}, or make use of additional artificial anomalies and ensemble methods, such as \textbf{CutPaste} \citep{li2021cutpaste}. Our model is trained from scratch and uses merely the provided training data, making for a less complex and more general method. The proposed \textbf{CPC-AD} approach is further outperformed by one autoencoding approach (\textbf{AE-SSIM}) and a partially contrastive approach (\textbf{DistAug}), but is on par with another autoencoding approach (\textbf{AE-L2}). Our proposed method outperforms the GAN-based approach (\textbf{AnoGAN}) \citep{Bergmann_2019_CVPR, schlegl2017unsupervised}. Interestingly, the CPC-AD model scores relatively well on textures, compared to similar models.

Nonetheless, although the quantitative results achieved with CPC-AD are not state-of-the-art, the model succeeds in generating accurate segmentation masks for most classes (\cref{fig:segmentation_results}). Even for classes with a low pixelwise AUROC score, such as pill, it can be seen that the created segmentation masks correctly highlight anomalous input regions, although there is some background noise. This corresponds with the comparatively high detection performance that the CPC-AD method achieves for this class (\cref{tab:auc_detection}). These results indicate that part of the low segmentation scores (compared to the detection scores) could be due to small spatial deviations from the ground truth. This effect might be exacerbated by the relatively low resolution of the segmentation masks that our patch-wise approach creates. Nonetheless, we argue that this resolution would be sufficient in practice to provide interpretable results for human inspection. Overall, CPC-AD provides a promising first step towards anomaly segmentation methods that are based on contrastive learning.

\section{Conclusion}
Overall, the CPC-AD model shows that contrastive learning can be used not just for anomaly detection, but also for anomaly segmentation. The proposed method performs well on the anomaly detection task, with competitive results for a majority of the data. Additionally 
the generated segmentation masks provide a promising first step towards anomaly segmentation methods that are based on contrastive losses.

\bibliography{main}
\bibliographystyle{icml2021}

\begin{table*}[t]
    \centering
    \caption{Anomaly segmentation -- pixelwise AUROC score on the MVTec-AD test-set per category. AE-SSIM and AE-L2 \citep{bergmann2020uninformed, liznerski2020explainable}, AnoGAN \citep{schlegl2017unsupervised, liznerski2020explainable}, DistAug in combination with KDE \citep{sohn2021learning} and CutPaste \citep{li2021cutpaste}, as well as our proposed CPC-AD model do not use a pre-trained feature extractor. FCDD \citep{liznerski2020explainable} and PaDiM \citep{defard2020padim} do make use of pre-training to improve their results (denoted with an *). }
        \begin{tabular}{llllllllll} \toprule
                   & \textbf{AE-SSIM} & \textbf{AE-L2} &\textbf{ AnoGAN} & \textbf{DistAug} & \textbf{CutPaste} & \textbf{FCDD*} & \textbf{PaDiM*} & \textbf{CPC-AD} &   \\ \midrule
        Bottle     & 0.93    & 0.86  & 0.86   & -     & 0.98  & 0.97 & 0.98  & 0.89   &   \\
        Cable      & 0.82    & 0.86  & 0.78   & -     & 0.9   & 0.9  & 0.97  & 0.84   &   \\
        Capsule    & 0.94    & 0.88  & 0.84   & -     & 0.97  & 0.93 & 0.99  & 0.72   &   \\
        Carpet     & 0.87    & 0.59  & 0.54   & -     & 0.98  & 0.96 & 0.99  & 0.74    &   \\
        Grid       & 0.94    & 0.9   & 0.58   & -     & 0.98  & 0.91 & 0.97  & 0.80   &   \\
        Hazelnut   & 0.97    & 0.95  & 0.87   & -     & 0.97  & 0.95 & 0.98  & 0.81   &   \\
        Leather    & 0.78    & 0.75  & 0.64   & -     & 1.0     & 0.98 & 0.99  & 0.94   &   \\
        Metal nut  & 0.89    & 0.86  & 0.76   & -     & 0.93  & 0.94 & 0.97  & 0.76   &   \\
        Pill       & 0.91    & 0.85  & 0.87   & -     & 0.96  & 0.81 & 0.96  & 0.77   &   \\
        Screw      & 0.96    & 0.96  & 0.8    & -     & 0.97  & 0.86 & 0.99  & 0.65   &   \\
        Tile       & 0.59    & 0.51  & 0.5    & -     & 0.91  & 0.91 & 0.94  & 0.82   &   \\
        Toothbrush & 0.92    & 0.93  & 0.9    & -     & 0.98  & 0.94 & 0.99  & 0.81   &   \\
        Transistor & 0.9     & 0.86  & 0.8    & -     & 0.93  & 0.88 & 0.98  & 0.90   &   \\
        Wood       & 0.73    & 0.73  & 0.62   & -     & 0.96  & 0.88 & 0.95  & 0.82   &   \\
        Zipper     & 0.88    & 0.77  & 0.78   & -     & 0.99  & 0.92 & 0.99  & 0.95   &   \\ \midrule
        Mean       & 0.87    & 0.82  & 0.74   & 0.90  & 0.96  & 0.92 & 0.98  & 0.82   &  \\ \midrule
        Texture mean & 0.78  & 0.70  & 0.58   & 0.83  & 0.97  & 0.93 & 0.97  & 0.83   & \\
        Object mean & 0.91   & 0.88  & 0.83   & 0.94  & 0.96  & 0.91 & 0.98  & 0.81    & \\\bottomrule
    \end{tabular}
    \label{tab:segmentation_auc}
\end{table*} 

\newpage \newpage

\appendix
\section{Additional Results}
\subsection{Anomaly Segmentation}

In \cref{tab:segmentation_auc}, we compare the anomaly segmentation performance of the proposed CPC-AD method against previously published works from peer-reviewed venues.

\end{document}